%% file: main_aacl.tex
\documentclass[11pt]{article}

\usepackage[final]{acl}

\usepackage{times}
\usepackage{latexsym}

\usepackage[T1]{fontenc}

\usepackage[utf8]{inputenc}

\usepackage{microtype}

\usepackage{inconsolata}

\usepackage{graphicx}

\usepackage{amsmath}
\usepackage{bbm}

\usepackage{algorithm}
\usepackage{algpseudocode}
\usepackage{graphicx}
\usepackage{wrapfig}

\newcommand{\algo}{FastVLM}

\title{FastVLM: Self-Speculative Decoding for Fast Vision-Language Model Inference}

\author{Divya Jyoti Bajpai \\
  Dept. of IEOR, IIT Bombay \\
  Mumbai, Maharashtra, India \\
  \texttt{divyajyoti.bajpai@iitb.ac.in} \And
  Manjesh Kumar Hanawal \\
  Dept. of IEOR, IIT Bombay \\
  Mumbai, Maharashtra, India \\
  \texttt{mhanawal@iitb.ac.in}}

\begin{document}
\maketitle
\input{chapters/abstract}

\input{chapters/introduction}

\input{chapters/related_works}

\input{chapters/motivation}
\input{chapters/methodology}

\input{chapters/experiments}

\input{chapters/conclusion}
\bibliography{custom}

\appendix

\input{chapters/appendix}

\end{document}

%% file: chapters/abstract.tex
\begin{abstract}
Vision-language Models (VLMs) have made significant strides in visual understanding and query response generation, but often face challenges of high computational cost and inference latency due to autoregressive decoding. In this work, we introduce an imitation-learning-based Self-Speculative Decoding (SSD) framework, named \algo{}, to address these limitations. Our approach employs a lightweight draft model for token generation in an autoregressive manner, while a full model verifies these tokens non-autoregressively. Accepted tokens proceed seamlessly, while rejected tokens are corrected by the full model and used to guide the draft model’s refinement. Through an imitation network, \algo{} enhances the draft model by integrating deeper-level insights from the full model's architecture. Also, it maintains the performance integrity of the full model while training the draft model, achieving a balance between efficiency and accuracy. Our method speeds up the inference process by $1.55-1.85\times$ as compared to the final layer with minimal loss in performance. The anonymized source code is available at \url{https://anonymous.4open.science/r/SSD-575E/README.md}.
\end{abstract}

%% file: chapters/introduction.tex
\section{Introduction}

Vision-language tasks have leveraged the benefits of  
 large Vision-Language models (VLMs) (e.g., BLIP-2 \cite{li2023blip} LLaVA \cite{liu2023visual}, LLaVA-1.5 \cite{liu2024improved}, etc.) to achieve state-of-the-art results. However, their high computation and memory requirements present challenges  \cite{samsi2023words, pope2023efficiently} to run them on resource-constrained devices, limiting their practical utility. This highlights the need to improve the inference speed of VLMs on resource-limited devices. 


One of the main efficiency bottlenecks in VLMs is autoregressive decoding. This decoding paradigm generates tokens sequentially, with each step requiring a separate invocation of the model to predict the next token given the previously generated tokens. Memory bandwidth limitations exacerbate this inefficiency, leading to overburdening computational resources and increased latency \cite{shazeer1911fast}. For instance, utilizing BLIP-2 with decoder FlanT5-xl \cite{chung2024scaling} for generating image captions can take roughly 13$\times$ longer with autoregressive decoding compared to a single forward pass for a sequence of the same length (averaged over COCO \cite{lin2014microsoft} validation images). This necessitates innovative decoding methods that can overcome these constraints.

Approaches such as pruning \cite{ michel2019sixteen, frantar2023sparsegpt,wang2022efficientvlm}, quantization \cite{zhang2020ternarybert, bai2020binarybert, frantar2022gptq, zhang2024sparsevlm} and knowledge distillation \cite{sanh2019distilbert, jiao2019tinybert, touvron2021training} have demonstrated effectiveness in deploying them on resource-constrained devices. However, these techniques require modifications to the model architecture, often involving parameter reductions in the backbone network, which can degrade overall performance.  

Recently, Speculative Decoding (SD) \cite{chen2023accelerating} has gained attention for speeding up autoregressive decoding. This method employs two models: a lightweight draft model that generates token predictions quickly in an autoregressive manner and a larger verification model that assesses the quality of the generated tokens in parallel. By incorporating a verification step, SD achieves the accuracy of the larger model while reducing the frequency of its invocation, resulting in substantial efficiency gains. 

\begin{figure}
    \centering
    \includegraphics[scale=0.5]{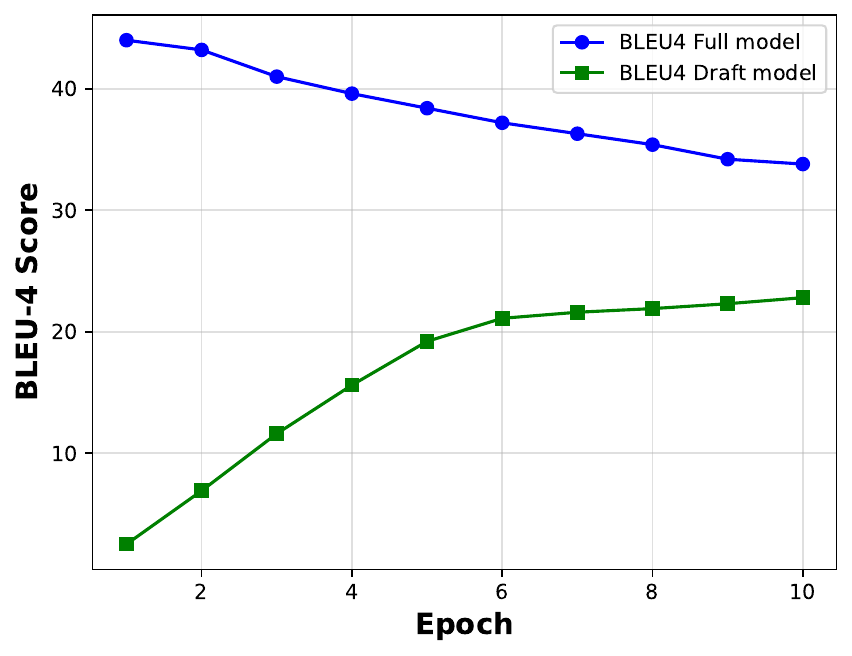}
    \caption{The effect on the final layer BLEU4 score due to combined training of the full model with the draft model.}
    \label{fig:loss_tradeoff}
\end{figure}
Despite its advantages, SD requires storing and executing two separate models, which can be impractical for devices with limited resources. To address this, we introduce Self-Speculative Decoding (SSD) in VLMs, originally proposed by Draft and Verify \cite{zhang2023draft} to reduce the computational burden of autoregressive language models. It utilizes the initial layers of the primary model as a lightweight draft model, while the full model verifies the generated tokens. This parameter-sharing design addresses storage and computational challenges, making SSD appealing for resource-constrained devices. SSD's efficiency heavily depends on the draft model’s quality: better draft predictions lead to higher token acceptance rates during verification, minimizing redundant decoding stages. However, adapting SSD to VLMs introduces significant challenges:

\textbf{Loss of deeper-layer information:} The draft model in SSD uses only the shallow layers of the decoder, excluding the rich multimodal features from deeper layers, which are critical for vision-language tasks. This diminishes the draft model’s accuracy and compromises overall efficiency.

\textbf{Shared objectives:} Sharing parameters between the draft and full model creates a trade-off between optimizing the draft model’s performance and preserving the performance of the full model. Figure \ref{fig:loss_tradeoff} illustrates this trade-off using BLEU-4 scores for the draft model and the full model on the BLIP-2 architecture, with the draft model comprising the first 12 layers of FlanT5-xl. Results are based on the validation split of the COCO dataset and highlight the difficulty in jointly optimizing these objectives.
These limitations necessitate a method that can improve the draft model's performance by accessing the deeper layer representations without losing the full model performance.

To overcome these limitations, we propose \algo{}, a novel SSD framework for VLMs that trains a lightweight network to imitate the behaviour of the full model, combining the features of imitation learning \cite{hussein2017imitation, zare2024survey} and knowledge distillation. It aligns the draft model's hidden representations with the deeper layers of the full model via cosine similarity as in \cite{fei2022deecap}, and further refines its predictions by distilling class probabilities. We refer to our model as an imitation-based network. 
The imitation network (IN) explicitly learns to recover the deeper-layer information excluded from the draft model. This is achieved by training the IN with a combination of ground-truth labels and deeper layer outputs while freezing the backbone parameters of the full model. This helps maintain its overall performance by decoupling the responsibilities of the draft model and enhancing its token generation capabilities without compromising the performance of the full model. 

Combining the draft model's shallow representations with the imitation network's deeper-layer approximations, \algo{} significantly improves token acceptance rates during verification (see figure \ref{fig:token_acceptance_rates}). This efficient integration ensures higher accuracy and minimizes computational overhead, offering a scalable solution. Also, since the full model and the draft model have shared parameters, the key-value (KV) cache, important for fast autoregressive decoding, can be reused for the deeper layers, where the KV cache can be reused during the verification step, significantly improving efficiency.


The main contributions of \algo{} are as follows:  
\begin{itemize} 
    \item We propose \algo{}, a novel end-to-end SSD framework to improve inference latency in VLMs using a lightweight imitation-learning-based draft model.  

    \item We develop a unique training approach for the draft model that effectively mimics deeper layers of the model, ensuring minimal information loss while maintaining the accuracy of the full model.


    \item By decoupling the responsibilities of the draft model, our method substantially increases token acceptance rates, leading to improved efficiency and fewer redundant computations. 

\item We demonstrate the effectiveness and efficiency of \algo{} through comprehensive experiments on MS-COCO \cite{lin2014microsoft}, NoCaps \cite{agrawal2019nocaps}, VisDial \cite{das2017visual}, MM-Vet \cite{yu2023mm}, and LLaVA-Wild \cite{liu2024improved} datasets on BLIP-2 and LLaVA-1.5 model, reducing the inference time by $1.55 \times-1.85\times$. Some qualitative examples on BLIP-2 and LLaVA-1.5 are given in Fig. \ref{fig:more_instances} and Fig. \ref{fig:llava_instance} respectively with discussion in Appendix \ref{sec:instance_discussion}.
\end{itemize}

%% file: chapters/related_works.tex
\section{Related works}

\textbf{DNN inference:}
Several works have designed systems specifically engineered for DNN inference. Some methods are Orca \cite{yu2022orca}, LightSeq \cite{wang2020lightseq}, PaLM inference \cite{pope2023efficiently}, TurboTransformers \cite{fang2021turbotransformers}, DeepSpeed Inference \cite{aminabadi2022deepspeed}, FlexGen \cite{sheng2023flexgen} etc. Despite these system optimizations, there are gaps in the careful co-design of algorithms and systems. This is necessary to fully harness the potential of hardware efficiency during DNN inference computation.

\textbf{Faster inference methods:}
Various model compression techniques have been explored to accelerate inference \cite{zhu2024survey}, including pruning \cite{frantar2023sparsegpt,wang2022efficientvlm, zhang2024sparsevlm}, quantization \cite{zhang2020ternarybert, bai2020binarybert, frantar2022gptq, wang2024q, bajpai2025survey}, and knowledge distillation \cite{sanh2019distilbert, jiao2019tinybert, touvron2021training}. These approaches statically modify the model structure.

In contrast, dynamic methods like early exiting \cite{bajpai2024capeen, bajpai2024ceebert, bajpai2024dadee, zhou2020bert, zhu2021leebert, moon2023early, huang2017multi, wolczyk2021zero, chataoui2023jointly} adaptively choose submodels during inference without altering the architecture. They leverage the fact that shallow layers often suffice for simpler inputs. While DeeCAP \cite{fei2022deecap} applies imitation learning for early exit tasks, we use imitation networks to decouple the draft model's role for speculative decoding.

\textbf{Speculative Decoding methods:} Speculative decoding (SD) \cite{chen2023accelerating, leviathan2023fast, li2024eagle, liu2023online, kim2024speculative} is a popular acceleration technique for autoregressive models. It leverages a lightweight draft model to generate multiple tokens sequentially, which are then verified in parallel by a larger, more accurate model. Recent works like Draft \& Verify \cite{zhang2023draft} and LayerSkip \cite{elhoushi2024layer} adopt Self-Speculative Decoding (SSD), where early decoder layers act as the draft model using layer-skipping strategies.

However, applying SSD directly to multimodal models is challenging, as deeper layers are often essential for capturing cross-modal interactions. Simply skipping layers and forwarding shallow outputs to the LM Head degrades performance. To address this, we propose an imitation-based draft model that learns to replicate deeper representations while remaining decoupled from the full model. It improves draft model performance, reduces backbone degradation, and improves token acceptance rates.

%% file: chapters/methodology.tex
\section{SSD Setup for VLMs}  




\begin{figure}
        \centering
        \includegraphics[scale=0.39]{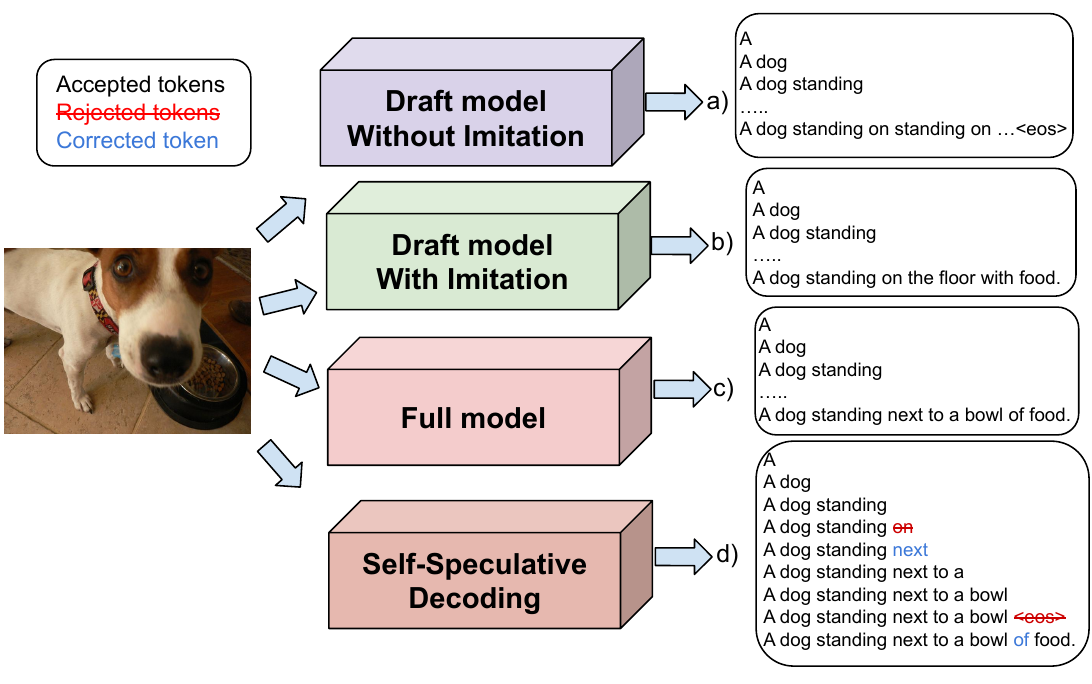}
        \caption{Inference from different model types: a) Draft without imitation; b) Draft with imitation; c) Full model; d) Self-Speculative Decoding.}
        \label{fig:inference_instance}
\end{figure}

We begin with a discussion on VLM architecture and the potential use of SSD for inference speedup. VLM  consists of an encoder and a decoder where the encoder part can be an image encoder, a combination of image encoder and image-grounded query generator \cite{li2023blip} etc., and the decoder is usually an LLM. The encoder extracts the multimodal features where the input is an image and a prompt. The extracted features are then passed to the decoder for image-grounded text generation. We denote the input image as $I$ and the input prompt as $T_0$. The encoder output is denoted as $z = E(I, T_0)$. $z$ is passed as an input to the decoder. Note that only the decoder is involved in autoregressive decoding, so we will mostly focus on the decoder of the VLM.  

The decoder consists of $L$ transformer layers with an embedding layer that maps the token indices to token embeddings $x_0$. The layer $l$ evolves embeddings output from its previous layer, $x_{l+1}^t = x_l^t+f_{l}(x_l^t)$ at every timestamp $t$ and a final Language Model (LM) Head that maps the embedding outputs of the last layer, $x_L$ to logits $o_L^t = g(x_L^t)$ where $t$ denotes that the output embeddings are for the $t$th token.

VLM decoders typically rely on an autoregressive decoding mechanism. In this framework, given a sequence of context tokens $T_0 = ( y^1, y^2, \ldots, y^t )$ and encoder output $z$, the task is to estimate the probability distribution \( P(y^{t+1} \mid y^1, y^2, \ldots, y^t, z) \) to predict the next token. The objective of the decoder is to correctly predict and estimate the output distribution's probability. The output $o_L^t$ is passed through a softmax layer and then the token with the highest probability is selected as the output, or alternatively, sampling can be performed based on the distribution to introduce variability and generate more diverse text outputs.  

The computational burden of autoregressive decoding is comparable to performing a forward pass over the entire sequence of tokens. Each time a new token is generated, all model parameters must pass through the computational pipeline on the hardware accelerator, such as a GPU or TPU. Consequently, the model's size, coupled with the available memory bandwidth, imposes a strict limit on decoding speed, leading to prolonged inference durations.

\subsection{Self-Speculative Decoding} 
This approach leverages two distinct models: a primary decoder model, which represents the full-capacity model, and a faster, smaller auxiliary model referred to as the `draft model'. In Self-Speculative Decoding (SSD), the first $n$ layers of the decoder model are considered the draft model.
The SSD process is divided into two distinct stages:  
 
   \textbf{1) Drafting Stage}: The draft model predicts a batch of \( d \) tentative tokens \( y^{t+1}, y^{t+2}, \ldots, y^{t+d} \) based on the context sequence \( y^1, y^2, \ldots, y^t \) autoregressively where $y_{i} = g(x_n^i)$ for $i\in \{t+1, \ldots, t+d\}$ and $x_n^i$ denotes the hidden state of the $n$th layer at $i$th decoding step. 
   
    \textbf{2) Verification Stage}: The full decoder is then used to verify the correctness of the draft tokens. In a single forward pass, the full decoder computes the probability for the drafted tokens and evaluates their alignment with the context. For any token, \( y^j \) that fails verification, the primary decoder's predicted token is used to replace \( y^j \), and the drafting process is restarted from \( y^{j+1} \) with tokens till $y^j$ as context.  

This dual-stage process significantly reduces the number of forward passes through the full decoder model while maintaining the quality of the generated output. However, a few technical challenges hinder its direct use in VLM.

\section{SSD issues in VLMs}
Below, we discuss the main issue in using SSD in VLMs and how introducing an imitation network on the draft model addresses the issue. 

\textbf{The Role of Deeper Layers in Complex Tasks:}  
This section underscores the role of deeper layers in VLMs. Our analysis highlights that multimodal tasks heavily depend on deeper layers of the backbone, where interactions between visual and textual elements necessitate rich, nuanced representations. These representations are formed in the deeper layers, making them indispensable for achieving convergence toward desired performance.

However, skipping these deeper layers, as in standard SSD, can significantly impact performance in intricate tasks. For example, as shown in Figure \ref{fig:inference_instance}, a draft model without imitation often falls into repetitive loops, predicting the same tokens without meaningful progression. To address this issue, we propose to use an imitation network. This enables the draft model to leverage deeper-layer insights, producing more accurate and coherent captions. Figure \ref{fig:inference_instance} is demonstrated using the BLIP-2 model with FlanT5-xl as the decoder, where our approach successfully bridges the gap between efficiency and complexity, enhancing overall system performance.

\begin{figure*}
    \centering
    \includegraphics[width=0.91\linewidth]{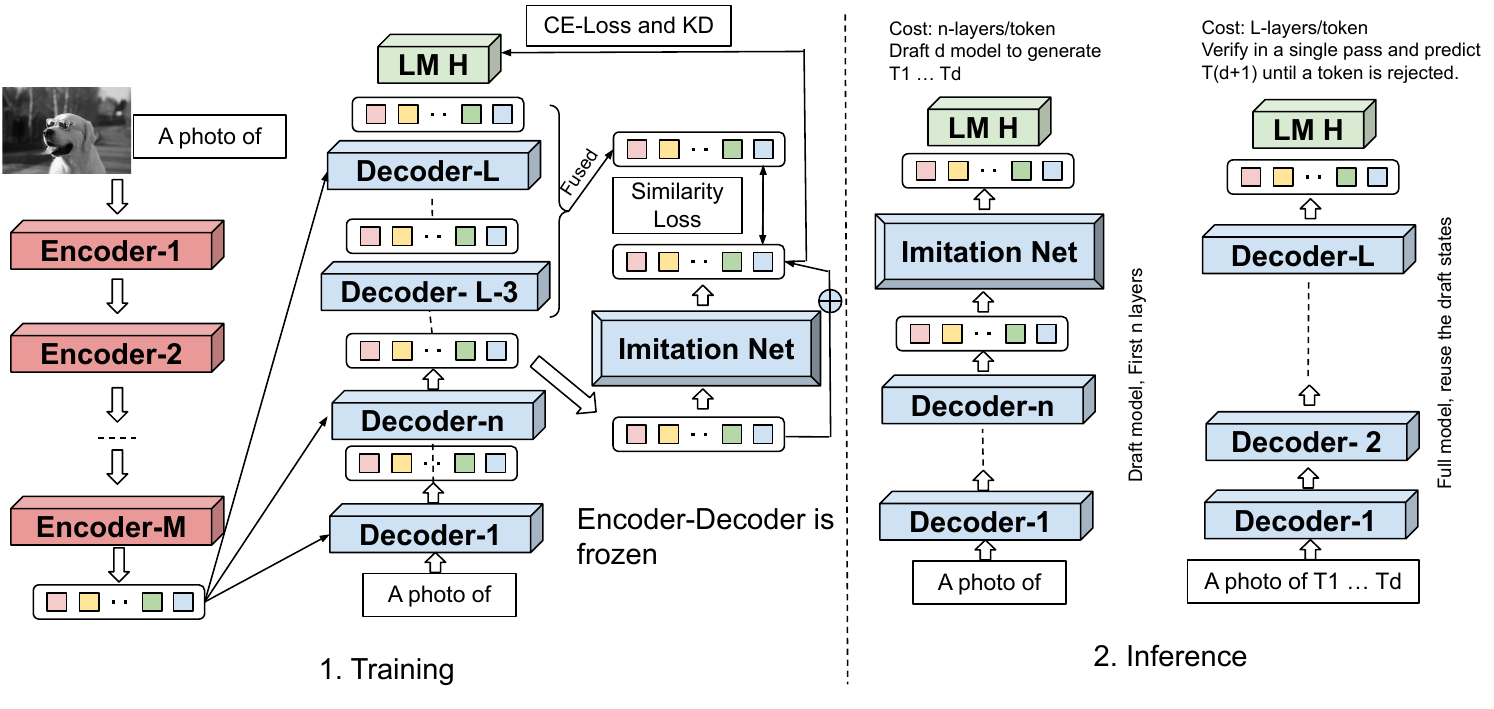}
    \caption{\textbf{Overview of our method.} \textbf{Left}: During training, the imitation network takes inputs from the \(n\)th decoder layer and learns to mimic deeper layer representations. Its output is passed to the LM Head, and training uses a combination of similarity loss, knowledge distillation, and cross-entropy loss. \textbf{Right}: At inference, the \(n\)th decoder layer and imitation network generate draft tokens, which are then verified by the full decoder.}
    \label{fig:full_SSD}
    \vspace{-0.5cm}
\end{figure*}

\textbf{Why Imitation Network?:}
Imitation learning was originally proposed for Reinforcement Learning methods to train an agent to mimic the behaviour of an expert by learning from demonstrations, without explicitly relying on a reward function. We use it to mimic the behaviour of the deeper layers. The reason for using an Imitation Network (IN) and not a specific LM Head for the draft model has multiple reasons: 1) Size: The size of the LM Head is $|x_l|\times |\mathcal{V}|$ where $x_l$ is the hidden state and $\mathcal{V}$ is the vocabulary of tokens which is huge. For instance, the hidden size of the FlanT5-xl is $2048$ and the vocab size is $32128$ which scales the LM Head size to $65M$ parameters! The IN is a much simpler network with only $39M$ parameters, reducing the overall training parameters by $40\%$. 2) Learning objectives: The objective of the LM Head is to map the representations given by the model to class probabilities, while the objective of the draft model is to mimic the deeper layer representations and then reuse the final layer LM Head, reducing the overall complexity. Also, the IN decouples the task of the draft model, preserving the final layer performance loss as shown in Figure \ref{fig:loss_tradeoff}.


Our approach with IN comes as a unified solution to multiple questions. 1) It preserves the full model accuracy by decoupling the task of the draft model. 2) It is lightweight and uses minimal parameters as compared to the backbone. 3) It has the ability to mimic deep-layer representations.

\section{Methodology}
We start with a pre-trained VLM backbone. The procedure is divided into three parts: 1) Fine-tuning the backbone. 2) Training the imitation network. 3) Then we discuss an efficient way to re-utilize the KV-Cache without recomputing it during the verification step. First, we start with the fine-tuning of the backbone.

\textbf{1) Fine-tuning the backbone:}
In this step, the model is fine-tuned for the final layer performance without any changes to the model architecture. The input is passed on to the full VLM and the final layer output is used to compute the loss given by
\begin{equation}
    \mathcal{L} = \mathcal{L}_{CE}(o_L^t, y^{t^*})
    \label{eqn:BBLoss}
\end{equation}
where $\mathcal{L}_{CE}$ is the cross-entropy loss, $y^{t^*}$ is the ground-truth token, and $o_L^t$ is the output logits of the decoder's final layer. After this step, the backbone parameters are frozen so that the optimality of the backbone is preserved.
 
\textbf{Deep Representation Imitation:}
From the analysis in Figure \ref{fig:inference_instance} suggests that even the prediction of simple tokens relies not only on the low-level features but also on high-level semantic information. However, directly accessing the deep representations is intractable since those are inaccessible until they feed forward the corresponding layers, which is not what we want. To bridge the gap, we approximate the uncomputed hidden states in deep layers using imitation learning. That is, we equip a lightweight network that is encouraged to predict the representation of the deeper layers based on the computed low-level representations. 

Formally, the input to the imitation network $\mathcal{I}$ is $x_n^t$, the output of the $n$th layer of the decoder at $t$th timestep. The imitation network is a much simpler model with fully connected layers. To find the architecture for the imitation network, we use a method similar to SelfxiT \cite{khademsohiselfxit} (see Appendix \ref{sec:imitation_network}).  Its objective is to provide a hidden representation such that it matches the deeper layer information. The output of the imitation network is denoted as ${x}_\mathcal{I}^t$. We utilize cosine-similarity loss to train the imitation network to mimic the deeper layers which is defined as:
\begin{equation}
    \text{Cos-Sim}({x}_\mathcal{I}^t, x_{fused}^t) = 1-\frac{{x}_\mathcal{I}^t\cdot x_{fused}^t}{||{x}_\mathcal{I}^t||\cdot||x_{fused}^t||}
\end{equation}
where $x_{fused}^t$ denotes the fused hidden states from deeper layers of the backbone and $||\cdot||$ denotes the L2 norm. The overall loss to train the imitation network is:
\begin{equation}
    \mathcal{L}_{imit}:= \text{Cos-Sim}+KL(p_\mathcal{I}^t, p_L^t)
    \label{eqn:ImitationLoss}
\end{equation}
where $KL$ is the KL divergence function defined as $KL(p_\mathcal{I}, p_L) = \sum_{v\in\mathcal{V}}p_\mathcal{I}(v)\log\frac{p_\mathcal{I}(v)}{p_n(v)}$. $p_\mathcal{} = softmax(g(\mathcal{I}(x_n^t)))$ and $p_L^t = softmax(o_L^t)$ and $\mathcal{V}$ is the vocabulary, i.e., the set of all the tokens.
We next discuss how we can get the fused hidden states from deeper layers.  

\textbf{Multi-level Representations Fusion:}
To perfectly guide the imitation network to learn the deeper level representation, we need to provide a good target hidden representation. For fusion, we take hidden states of the last four layers i.e. $x_L^t, x_{L-1}^t, x_{L-2}^t \text { and } x_{L-3}^t$. The number of hidden representations that are good to fuse is explored in \cite{devlin2018bert, horne2020grubert}  observing that the best performance is when the last three or four layers are fused together. The three methods for fusion are:

\textbf{i) Averaging:} All the hidden representations in different layers are averaged directly.

\textbf{ii) Concatenation:} All the hidden representations are concatenated in the sequence dimension and then fed into a linear transformation layer to obtain a final compressed representation.

\textbf{iii) Attention-pooling:} Utilizes the weighted projection of all the hidden representations as the integrated information. The attention weights are computed with the last hidden representation as the query and hold a certain robustness to noise.

\begin{algorithm}[ht]
   \caption{Self-Speculative Imitation Decoding (greedy)}
   \label{alg:SSD}
\begin{algorithmic}[1]
   \State {\bfseries Input:} $z = E(I,T_0), y^1, y^2 \cdots y^t$ could be input prompt to decoder or just the $<sos>$ token, target sequence length $T$; max draft tokens to generate $d$.
   \State $i \gets t$
   \While{$i<T$ or $<eos>$ is generated}
   \For{$j\gets i, i+1, \ldots, i+d$}
   \State $p_\mathcal{I}(y|z, y^1, ..,y^j)\gets \mathcal{S} (g(\mathcal{I}(x_n^{j+1})))$
   \State $y^{j+1}\gets \arg \max p_\mathcal{I}(y|z, y^1, \ldots,y^j)$ 
   \EndFor
   \For{$i\gets i,\ldots,j$}
   \If{$y^{i+1}\neq \arg\max p_L(y|z, y^1, \ldots, y^i)$}
   \State $y^{i+1}\gets \arg\max p_L(y|z, y^1, ., y^i)$
   \State Break
   \EndIf
   \State $i\gets i+1$
   \State If all draft accepted, then generate next token $y^{i+1}\gets \arg\max p_L(y|z, y^1, \ldots, y^i)$ and $i\gets i+1$
   \EndFor
   \EndWhile
   \State {\bfseries return} $y^1, y^2 \ldots y^T$
\end{algorithmic}
\end{algorithm}

These methods give the $x_{fused}^t$ representations consisting of the fine-grained deeper layer knowledge. This guides the imitation network to generate hidden representations such that high-level information available at the deeper layers is not lost. 

\textbf{2) Training Draft Model:}
This section details the training procedure of the draft model. Draft model in our case consists of the first $n$ layers of the backbone and the imitation model. As the backbone is frozen, the only trainable component left is the imitation network. The output of the $n$th layer is passed as an input to the imitation network that is used to mimic the behaviour of the remaining $L-n$ layers of the backbone. The overall loss function used for training is:
\begin{equation}
    \mathcal{L}_{draft}:= \mathcal{L}_{imit}+\mathcal{L}_{CE}(o_\mathcal{I}^t, y^{t^{*}}) 
\end{equation}

where $\mathcal{L}_{CE}$ is define in Eq.~\ref{eqn:BBLoss} and $\mathcal{L}_{imit}$ is given in Eq.~\ref{eqn:ImitationLoss}. The left part of Figure \ref{fig:full_SSD} provides a visual representation of the training process.

\begin{table*}[]
\centering
\small
\begin{tabular}{lcccccccc}
\hline
Models          & n  & BLEU1 & BLEU4 & CIDEr & SPICE & METEOR & ROUGE-L                          & Spd. \\ \hline
\multicolumn{9}{c}{\textit{BLIP2-FlanT5}}                                                               \\ \hline
BLIP2-FlanT5      & 24 & 84.1  & 44.0    & 144.6 & 24.6  & 30.9   & 61.8                             & 1.00$\times$    \\ \hline
Draft w/o Im    & 12 & 05.2   & 00.1   & 01.7   & 0.00     & 00.4    & 02.5                              & 2.00$\times$    \\
Draft w Im      & 12 & 65.5  & 18.9  & 86.3  & 14.1  & 17.6   & 47.5                             & 2.00$\times$    \\
Draft \& Verify & 12 & 74.1  & 32.9  & 127.8 & 21.7  & 28.9   & 55.7                             & 1.27$\times$ \\
LayerSkip       & 12 & 75.6  & 33.4  & 129.1 & 21.9  & 29.5   & 56.5                             & 1.25$\times$ \\ \hline
Our          & 12 & 82.4  & 42.5  & 139.7 & 23.6  & 29.7   & 59.9                             & 1.39$\times$ \\
Our+CS       & 12 & 83.2  & 43.3  & 141.9 & 24.1  & 30.4   & 60.6                             & 1.53$\times$ \\
Our+CS+KD    & 12 & 83.8  & 43.6  & 142.6 & 24.3  & 30.7   & 61.1                             & 1.61$\times$ \\ \hline
\multicolumn{9}{c}{\textit{BLIP2-OPT}}                                                                  \\ \hline
BLIP2-OPT       & 32 & 83.5  & 43.7  & 143.1 & 24.4  & 30.9   & 61.5                             & 1.00$\times$    \\ \hline
Draft w/o Im    & 15 & 07.9   & 01.3   & 02.5   & 00.1   & 00.8    & 05.2 & 2.13$\times$    \\
Draft w Im      & 15 & 65.8  & 19.5  & 91.7  & 15.4  & 18.1   & 47.5                             & 2.13$\times$    \\
Draft \& Verify & 15 & 72.7  & 31.4  & 125.2 & 20.9  & 28.1   & 54.4                             & 1.32$\times$ \\
LayerSkip       & 15 & 73.5  & 31.9  & 127.0   & 21.2  & 28.5   & 55.9                             & 1.39$\times$ \\ \hline
Our          & 15 & 81.8  & 42.5  & 139.8 & 23.6  & 29.5   & 59.5                             & 1.49$\times$ \\
Our+CS       & 15 & \underline{82.3}  & \underline{43.0}  & \underline{142.4} & \underline{23.9}  & \underline{30.0}     & \underline{60.3}                             & \underline{1.71}$\times$ \\
Our+CS+KD    & 15 & \textbf{82.9}  & \textbf{43.4}    & \textbf{142.5} & \textbf{24.3}    & \textbf{30.5}   & \textbf{61.2}                             & \textbf{1.75}$\times$ \\ \hline
\end{tabular}
\caption{Results on the test split of COCO dataset, Spd. shows the average speedup in the inference. Im in the table denotes the imitation network. CS denotes cosine similarity loss and KD denotes Knowledge Distillation.}
\label{tab: coco_res}
\end{table*}

\textbf{Inference:}
In Algorithm \ref{alg:SSD}, we outline the inference process where the input is $z = E(I, T_0)$, the encoder output, input prompt, target sequence length and number of draft tokens to be generated before verification. The inference process could be divided into two stages: 1) Draft token generation (lines 4-7): The first step is to generate the set of $d$ draft tokens using the imitation network and the first $n$ layers of the backbone. 2) Verification (lines 8-14): where the model verifies each generated token and if one of these tokens is not the same as the final layer's token the verification process is stopped with the corrected token appended to the caption and the process returns to the drafting stage. This process is repeated until the target sequence length is achieved or the end-of-sentence $<eos>$ token is predicted. For a visual representation of inference process see right side of Figure \ref{fig:full_SSD}.

\textbf{Choice of $d$ and $n$:} The value of $d$ i.e., the number of draft tokens to be generated before a verification step effects the number of acceptances. A higher value of $d$ can lead to more wasted generation from the draft model if there are rejections while a smaller value of $d$ can lead to more number of full model calls. We perform an ablation study in Figure \ref{fig:d_vs_speed}. However, from Figure \ref{fig:token_acceptance_rates} and Appendix \ref{sec:impact_of_d}, we observe that the token acceptance rate increases with a larger context length. We use this information to choose a dynamic value of $d$ in terms of context length $t$ where the value $d$ increases with increase in $t$ (see Appendix \ref{sec:impact_of_d}). For the choice of $n$, i.e., the layer up to which the model is treated as draft model can be chosen based on user-requirements, however, we provide a method based on a reward function (see Appendix \ref{sec:choice_of_n}).


\textbf{KV cache:} Key-Value cache is an important part of efficient generation in auto-regressive decoding that allows not to recompute the KV pairs at each step of the decoding. As the draft model and the verification model come from the same backbone using the same order of layers. Also, the first $n$ layers are shared for both methods, hence, the KV cache of the first $n$ layers is already computed, so we are able to maintain a single KV cache for the draft and verify step, saving memory and latency.


%% file: chapters/experiments.tex
\section{Experiments}
We evaluate our method on image captioning and visual dialogue using BLIP-2, which enables strong performance and flexible decoder selection. For more complex vision-language tasks requiring multi-step reasoning, we use LLaVA-1.5-7B~\cite{liu2024improved}, built on a vision encoder and LLaMA-2~\cite{touvron2023llama} decoder. This setup covers a range of SOTA models across VL tasks. Additionally, we include results on a task-tuned CLIP–LLaMA encoder-decoder model in Appendix~\ref{sec:encoder_decoder}.


\textbf{Datasets:} We evaluate our method on image captioning (COCO~\cite{lin2014microsoft}, NoCaps~\cite{agrawal2019nocaps}), visual dialog (VisDial~\cite{das2017visual}), and complex multimodal reasoning (MM-Vet~\cite{yu2023mm}, LLaVA-Wild~\cite{liu2024improved}). Captioning metrics include BLEU-4, METEOR, CIDEr, and SPICE; VisDial is evaluated using MRR; and MM-Vet and LLaVA-Wild follow the LLM-as-a-judge protocol~\cite{liu2024improved}. Speedup is reported as average per-token inference time reduction~\cite{zhang2023draft, elhoushi2024layer}. Results on VQAv2~\cite{goyal2017making} appear in Appendix~\ref{sec:vqa}.

\textbf{Training:} We utilize two variants of the BLIP-2 model where the encoder is kept the same (ViT-g/14 \cite{alexey2020image}) and the LLMs used are OPT\textsubscript{2.7B} \cite{zhang2022opt} and the FlanT5-xl \cite{chung2024scaling} models. For the LLaVA model, we have used LLaMA-2 \cite{touvron2023llama} as the decoder model.
We have two steps during training: 1) Finetuning the backbone: Here, the full backbone is fine-tuned. It is the usual fine-tuning step (not required for the LLaVA model).
2) Imitation Network tuning: During this step, the model weights are frozen, and we attach an imitation network to the $n$th layer of the decoder. The architecture of the imitation network consists of two fully connected layers and one layer norm. In this step, the model is trained for 20 epochs. More hyperparameter details can be found in Table \ref{tab: training_hyper_1}, \ref{tab: training_hyper_2} and \ref{tab: training_hyper_3} in the Appendix. We show the advantages of freezing the backbone parameters in Figure \ref{fig:Freezed_training} and a discussion in Appendix \ref{sec:impact_of_freezing}.

\begin{table*}[]
\centering
\small
\begin{tabular}{l|ccccccc|c|c}
\hline
Models         & \multicolumn{7}{c|}{MM-Vet}                     & LLaVA-W & Spd. \\
                & Rec  & OCR  & Know & Gen  & Spat & Math & Total &            &      \\ \hline
LLaVA-7B        & 33.1 & 20.0   & 18.9 & 20.1 & 25.8 & 05.1  & 28.0    & 64.8       & 1.00$\times$    \\ \hline
Draft w/o Im    & 00.6  & 00.1  & 00.0    & 00.4  & 00.7  & 00.0    & 00.3   & 04.6        & 2.00$\times$    \\
Draft w Im      & 21.3 & 10.8 & 11.5 & 12.2 & 12.9 & 01.7  & 15.8  & 45.2       & 2.00$\times$    \\
Draft \& Verify & 26.4 & 15.2 & 14.1 & 16.0   & 17.4 & 03.5  & 22.4  & 53.9       & 1.64$\times$ \\
LayerSkip       & 28.2 & 16.5 & 15.3 & 17.8 & 20.5 & 04.1  & 25.7  & 58.4       & 1.71$\times$ \\ \hline
{Ours}            & \textbf{32.8} & \textbf{19.5} & \textbf{18.7} & \textbf{19.9} & \textbf{25.7} & \textbf{05.0}    & \textbf{27.8}  & \textbf{64.1}       & \textbf{1.85}$\times$ \\ \hline
\end{tabular}
\caption{Results of the LLaVA-1.5-7B (LLaMA-2) model on MM-Vet and LLaVA-Wild datasets.}
\label{tab:llava_res}
\end{table*}

For the architecture of the imitation network, we considered options such as fully connected layers (up to 2) of the same size as the backbone, and transformer layers (up to one) and used Neural Architecture Search (NAS) to obtain the best architecture (see Appendix \ref{sec:imitation_network}). We observed that the combination of two fully connected layers performed better as compared to one transformer layer. Next, the fusion strategy used was the concatenation of the hidden states as it shows better performance (see Table \ref{tab:fusion_res}). Note that the loss in performance is small across fusion strategies. 

To find the optimal layer for the draft model we use the method provided in section \ref{sec:choice_of_n}. We provide an ablation study on different values of $n$ in Appendix \ref{sec:lambda}. We also provide details on the number of draft tokens to be generated before the verification step in Appendix \ref{sec:impact_of_d}.


\textbf{Baselines:}
\textbf{1) Full Model:} Standard BLIP-2 inference using OPT and Flan-T5-xl decoders.
\textbf{2) Draft (w/o Imitation):} Output from the $n$th layer is directly passed to the LM head, skipping remaining layers.
\textbf{3) Draft (w/ Imitation):} The $n$th layer output is refined through an imitation network before passing to the LM head.
\textbf{4) LayerSkip:} We implement LayerSkip as proposed, adapting it to vision-language models.
\textbf{5) Draft and Verify:} Extended to VLMs; unlike LayerSkip, it halts draft token generation when confidence drops below a threshold.
\textbf{6) Ours:} Our method with imitation trained using (i) cross-entropy loss, (ii) +cosine similarity (Our+CS), and (iii) +cosine similarity \& knowledge distillation (Our+CS+KD). 


\subsection{Results}
In Tables \ref{tab: coco_res} and \ref{tab: nocaps_res}, we present results on the Karpathy COCO test split, VisDial, and NoCaps datasets, where our method consistently outperforms baselines. We observe a similar behaviour in Table \ref{tab:llava_res}, where we provide results on the MM-Vet and LLaVA-Wild datasets. The draft model without the imitation network (IN) is analogous to skipping the last $L-n$ layers and directly passing the $n$th layer's representations to the LM Head, resulting in poor performance due to the mismatch with the LM Head's training. In contrast, the draft model with the IN demonstrates the IN's impact by significantly improving draft performance. The performance drop in Draft \& Verify and LayerSkip stems from their joint training of the draft and final layers, which compromises final-layer performance.

Our method achieves superior results due to: (1) Decoupling the draft model, which avoids competing objectives by using a small imitation network and freezing backbone parameters post-fine-tuning, preserving final-layer performance; and (2) Deeper layer mimicking, where the imitation network learns patterns from deeper layers, enhancing token acceptance and overall speedup. The contribution of each loss component is also shown in the tables, highlighting their role in improving performance. The better speedup in our method comes from factors such as lightweight imitation network, better performance of the draft model and mimicking deep-layer representation improves token acceptance rate. We can also observe the importance of different components in the loss function used to train the backbone. Observe that directly using cross-entropy loss, the speedup drops significantly as the draft model is not well trained. This shows that adding the Cosine Similarity and Knowledge Distillation loss makes the model more efficient.



%% file: chapters/conclusion.tex
\section{Conclusion}
We proposed an SSD method that can speed up the inference process in Vision-Language Models (VLMs). Our method introduces a lightweight imitation network trained to mimic the behaviour of deeper layers. We also provide a unique training approach such that the imitation network can learn to transform the input hidden representation into a representation that is a better approximation of the deeper layers. The full model performance is also preserved by decoupling the task of improving the draft model performance and the final layer performance. Experiments on multiple vision-language tasks and decoders further support our claims.

\section{Limitations}\label{sec:limitations}
Our method builds upon self-speculative decoding (SSD), wherein a draft model—formed by a subset of the full model and an additional imitation network—generates preliminary outputs that are then verified or corrected by the full model. This setup yields substantial speedups for long-form generation tasks by minimizing the number of full model invocations. However, in tasks such as Visual Question Answering (VQA), where the output typically consists of only one or two tokens, the potential for acceleration diminishes. Since the draft model generates only a small number of tokens before requiring verification, the frequency of full model calls becomes comparable to that of the draft model. This reduces the relative computational gain achievable through SSD. Despite this limitation, our method still offers improved efficiency over using the full model alone due to the shared parameters and architectural overlap between the draft and full models—an efficiency advantage that is not inherent to traditional speculative decoding, where the draft and final models are entirely separate.

\section{Impact Statement}\label{sec:impact_statement}

This paper presents a method to improve inference latency in Vision Language Models. The work can potentially improve the inference of large machine language models.

As usual, these can improve societal conditions, but of course, as with any technology, specific deployments need care. However, this is outside of the scope of the present work, which is aimed at improving the basic machine-learning models.

\section*{Acknowledgements}
Divya Jyoti Bajpai is supported by the Prime Minister’s Research Fellowship (PMRF), Govt. of India.  Manjesh K. Hanawal thanks funding support from SERB, Govt. of India, through the Core Research Grant (CRG/2022/008807) and MATRICS grant (MTR/2021/000645), and DST-Inria Targeted Programme.  We also thank funding support from Amazon IIT-Bombay AI-ML Initiative (AIAIMLI).

%% file: chapters/appendix.tex
\newpage
\section{Appendix}

\begin{table*}[]
\small
\centering
\begin{tabular}{lcccccccccc}
\hline
\textbf{Model}   & \textbf{VisDial} & \multicolumn{8}{c}{\textbf{NoCaps Zero-shot}}                                                                                       &      \\
                 &                  & \multicolumn{2}{c}{in-domain} & \multicolumn{2}{c}{near-domain} & \multicolumn{2}{c}{out-domain} & \multicolumn{2}{c}{full-dataset} & Spd. \\
                 & MRR              & C              & S            & C               & S             & C              & S             & C               & S              &      \\ \hline
BLIP-2-FlanT5-xl & 45.8             & 123.7          & 16.3         & 120.7           & 16.0            & 124.5          & 15.1          & 121.6           & 15.8           & 1.00$\times$    \\
Draft w/o imi    & 00.3              & 02.4            & 0.00            & 02.1             & 0.00             & 02.0             & 0.00             & 02.3             & 0.00              & 2.00$\times$    \\
Draft w imi      & 23.1             & 60.5           & 09.7          & 58.6            & 09.1           & 61.0             & 09.8           & 60.3            & 09.6            & 2.00$\times$    \\
Draft \& Verify  & 31.5             & 108.1          & 13.8         & 107.5           & 13.4          & 109.5          & 13.0            & 107.9           & 12.8           & 1.38$\times$ \\
LayerSkip        & 33.2             & 109.9          & 14.1         & 108.2           & 13.7          & 110.0            & 13.2          & 108.3           & 13.0             & 1.45$\times$ \\ \hline
Ours             & \textbf{43.9}             & \textbf{122.0}            & \textbf{15.9}         & \textbf{119.4}           & \textbf{15.5}          & \textbf{123.1}          & \textbf{14.8}          & \textbf{119.8}           & \textbf{15.3}           & \textbf{1.55}$\times$ \\ \hline
\end{tabular}
\caption{Results on the VisDial and the NoCaps dataset. For the NoCaps dataset, a zero-shot evaluation is performed using a model trained on the COCO dataset.}
\label{tab: nocaps_res}
\end{table*}

\begin{figure*}
        \centering
        \includegraphics[scale=0.39]{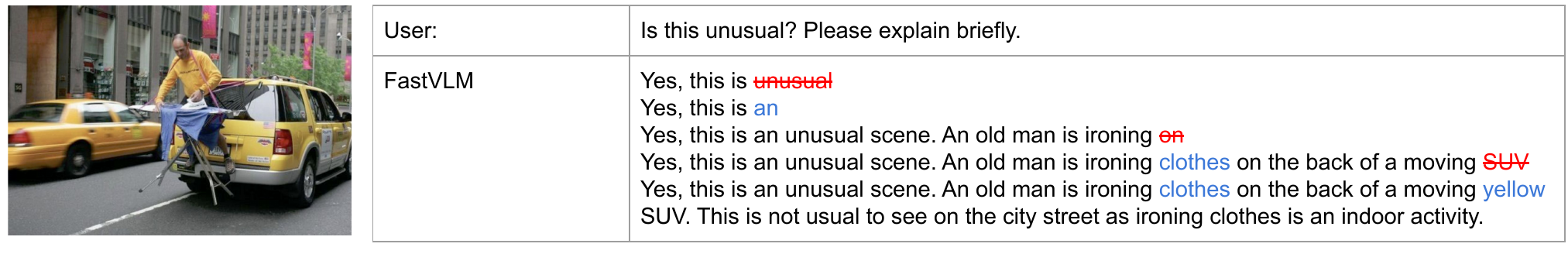}
        \caption{An instance of our method applied to LLaVA-1.5 on a long-context example.}
        \label{fig:llava_instance}
\end{figure*}

\begin{table*}[]
\centering
\small
\begin{tabular}{lcccccccc}
\hline
Baselines          & n  & BLEU1 & BLEU4 & CIDEr & SPICE & METEOR & ROUGE-L & Spd. \\ \hline
CLIP-LLAMA       & 32 & 81.6  & 41.5  & 138.5 & 22.8  & 29.9   & 60.2    & 1.00$\times$     \\
Draft w/o Im    & 13 & 07.9  & 01.4   & 03.1   & 00.0   & 00.7    & 07.1     & 2.46$\times$    \\
Draft w Im      & 13 & 52.3  & 13.7  & 59.2  & 15.9  & 20.6   & 47.0      & 2.46$\times$     \\
Draft \& Verify & 13 & 75.3  & 27.2  & 109.5 & 18.1  & 25.8   & 53.7    & 1.49$\times$  \\
LayerSkip       & 13 & 76.1  & 27.4  & 111.8 & 18.5  & 26.0     & 55.2    & 1.45$\times$  \\ \hline
Our             & 13 & \textbf{80.3}  & \textbf{40.7}  & \textbf{136.5} & \textbf{22.1}  & \textbf{29.3}   & \textbf{59.0}    & \textbf{1.77}$\times$  \\ \hline
\end{tabular}
\caption{Results on the COCO test split where the encoder-decoder are CLIP and Llama model respectively.}
\label{tab:res_encoder_decoder}
\end{table*}

\begin{table}[]
\centering
\small
\begin{tabular}{lccccc}
\hline
\textbf{Fusion-Strategy} & n  & B4 & CIDEr & MR & Spd. \\ \hline
\textbf{Average}         & 15 & 43.1  & 141.0   & 30.3   & \textbf{1.68}$\times$ \\
\textbf{Concatenation}   & 15 & \textbf{43.6}  & \textbf{142.6} & \textbf{30.7}   & 1.67$\times$ \\
\textbf{Attention-pool}  & 15 & 43.4  & 141.8 & 30.7   & 1.59$\times$ \\ \hline
\end{tabular}
\caption{Results showing the impact of different fusion strategies for better knowledge transfer from deeper layers to draft model on the COCO dataset.}
\label{tab:fusion_res}
\end{table}

\begin{table*}[]
\centering
\small
\begin{tabular}{ccccccc}
\hline
\textbf{$\lambda$} &\textbf{n}  & BLEU4 & CIDEr & METEOR & SPICE & Spd. \\ \hline
$1/10$  & \textbf{5}  & 42.4  & 138.7 & 29.8   & 23.9  & 1.88$\times$  \\
$1/15$ & \textbf{8}  & 43.2  & 140.4 & 30.2   & 24.1  & 1.75$\times$ \\
$1/L$ & \textbf{12} & 43.6  & 142.6 & 30.7   & 24.3  & 1.61$\times$ \\
$1/30$ & \textbf{16} & 43.7  & 143.1 & 30.8   & 24.4  & 1.42$\times$ \\
$1/50$ &  \textbf{21} & 43.9  & 144.0   & 30.9   & 24.6  & 1.17$\times$ \\ \hline
\end{tabular}
\caption{Impact of the value of $\lambda$ on the draft model size and the overall model performance.}
\label{tab:lambda_values}
\end{table*}
\subsection{Results using the encoder-decoder backbone}\label{sec:encoder_decoder}
In Table \ref{tab:res_encoder_decoder}, we provide the results on the encoder-decoder backbone where the encoder used is the CLIP-base model and the decoder is the LLaMA-7B model. The model was trained under the same hyperparameter setting. The results observed align to the findings on the BLIP-2 model. The input prompt in this case is `Describe the image' which is passed to the CLIP model that generates the prompt-based embeddings. These embeddings are then transformed by a linear layer to be given as input to the decoder. LLAMA model then utilizes the SSD approach to draft the tokens using the draft model with imitation and then the full LLAMA backbone is used for verification.

\subsection{Results on the VQA dataset}\label{sec:vqa}
In Table \ref{tab: vqa_res}, we provide the results on the VQA dataset. Observe that over the same setup, the speedup in this case is quite low. This happens due to the small answer length in the VQA tasks. Instead of keeping the draft token generated at once to a small value i.e., 5, we have observed that as there were many yes/no type answers and one-word answers, the speedup was not as significant as in tasks where the generated texts has a large number of words. However, the accuracy behaviour is similar to the other tasks. Note that all the baselines also have a lower speedup for similar reasons.

\begin{figure}
    \centering
    \includegraphics[scale=0.5]{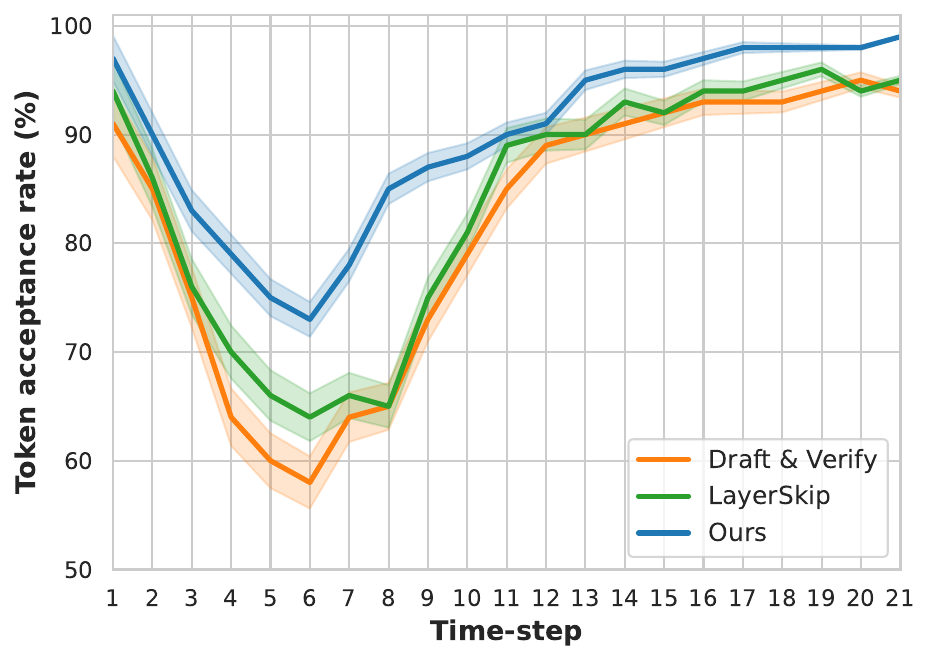}
    \caption{The token acceptance rates across different time-steps, i.e., at the $t$th token generation when $t-1$ tokens are used as a context.}
    \label{fig:token_acceptance_rates}
\end{figure}

\subsection{Choice of $d$}\label{sec:impact_of_d}
\begin{figure}
    \centering
    \includegraphics[scale=0.5]{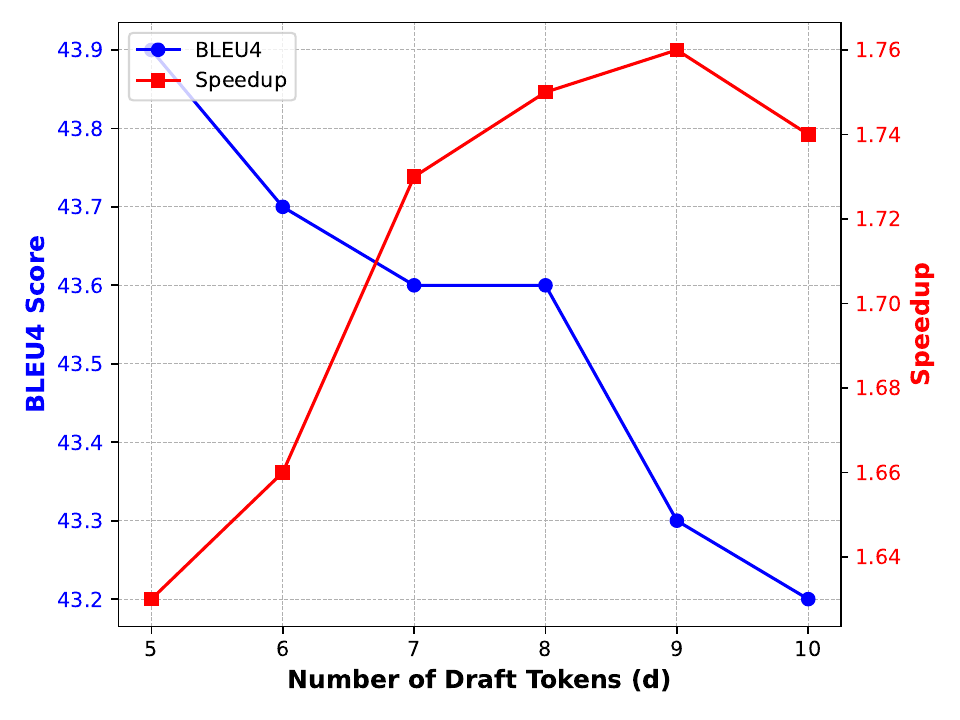}
    \caption{The impact of speedup and BLEU4 score over the change in the values of the number of draft tokens to be generated before a verification step.}
    \label{fig:d_vs_speed}
\end{figure}

The choice of \( d \), representing the number of draft tokens generated before a verification step, is a critical hyperparameter. Selecting an appropriate value for \( d \) can significantly reduce redundant generations from the draft model and minimize the number of calls to the full model. To evaluate the effect of a static choice of \( d \) across time steps, we conduct an ablation study, which is illustrated in Figure \ref{fig:d_vs_speed}.

However, as shown in Figure \ref{fig:token_acceptance_rates}, we observe a non-linear behavior in the token acceptance rate as the context length increases. Initially, the acceptance rate decreases, but it later increases and saturates at a higher value. This indicates that tokens generated with a larger context length are more likely to be accepted during the verification step. Based on this observation, we opt for a dynamic value of \( d \) that gradually increases as the context length grows. Specifically, the value of \( d \) at the \( t \)-th time step is defined as:
\begin{equation}
d(t) = d(0) + \left\lfloor 10 \cdot \frac{1}{1 + e^{-0.01(t - \frac{N}{2})}} \right\rfloor
\end{equation}
where we set \( d(0) = 5 \), ensuring that \( d(t) \) remains within the range \( \{5, 6, \ldots, 15\} \), $N$ denote the maximum sequence length of the model and $\left\lfloor\cdot\right\rfloor$ denotes the floor function. By allowing \( d \) to increase with the context length, we reduce the number of calls to the full model. However, setting a higher static value for \( d \) without considering context length could lead to an increase in rejections of generated tokens by the draft model, resulting in efficiency losses. On the other hand, increasing \( d \) as \( t \) progresses allows for greater draft model reliability with longer context lengths, thus maintaining efficiency without sacrificing performance.

\subsection{Choice of $n$}\label{sec:choice_of_n}
In SSD, the first $n$ layers of the backbone are used as the draft model. The choice of $n$ models the effectiveness versus efficiency trade-off. The effectiveness comes from a higher token acceptance rate during verification where a larger $n$ can improve it. While efficiency decreases as more computational cost is required to get the draft tokens. Thus, the choice of $n$ is important to achieve the overall performance of the SSD method. Also, for the candidate set of layers, we take $n>L/2$ so that the draft model is not computationally heavy.   

We choose the value of $n$ by maximizing the reward function based on the confidence at an intermediate layer $l$ i.e., $C_l:=\max_{v\in\mathcal{V}} softmax(g(x_l))$ and the computational cost to predict the token from the $l$th layer set as $\lambda\cdot l$. We choose $\lambda = 1/L$, which denotes the per layer computational cost of the decoder. If $\mathcal{D}$ represents the distribution of the incoming images, the total expected reward for layer $l$ is defined as $R(I, l) = E_{I\sim D}[r(I, l)]$ where $r(I, l) = C_l-\lambda\cdot l$. The layer that is chosen is defined as:
   $ n = \arg\max_{l\in[L/2]}E_{I\sim D}[r(I, l)]$
where $[L/2] = \{1, 2,\ldots, L/2\} $. The empirical reward could be defined as $\hat{R}(I, l) = \frac{\sum_{i=1}^{|\mathcal{D}|}r(I, l)}{|\mathcal{D}|}$ is used to find the value of $n$ over the validation split of the dataset. Note that the reward is such that it will maximize confidence while minimizing the computational aspects. Also, it will restrict the draft model to a layer after which there is very little improvement in confidence.

\subsection{Ablation over $\lambda$}\label{sec:lambda}
In table \ref{tab:lambda_values}, we show the impact of the $\lambda$ values over the full model performance and the draft model size. A higher value of $\lambda $ restricts the draft model to a lower value of $n$ as the reward function has a higher penalty for going deeper to the backbone and vice versa. Also, to show the impact of the deeper layers, if chosen as a draft model, we remove the restriction of $n\leq L/2$ only for the results in Table \ref{tab:lambda_values}. This is done to show how the model performs in terms of efficiency as deeper layers are chosen for the draft model. One interesting observation is that as we increase $\lambda$, the value of $n$ decreases, the model still does not take a huge hit in terms of performance that comes due to the imitation network, as it has to take input from an intermediate layer. It learns to map the representations at any intermediate layer to representations similar to the deeper layers. This makes our method robust to different values of $n$.

\subsection{Impact of freezing the backbone}\label{sec:impact_of_freezing}
\begin{figure}
    \centering
    \includegraphics[scale = 0.5]{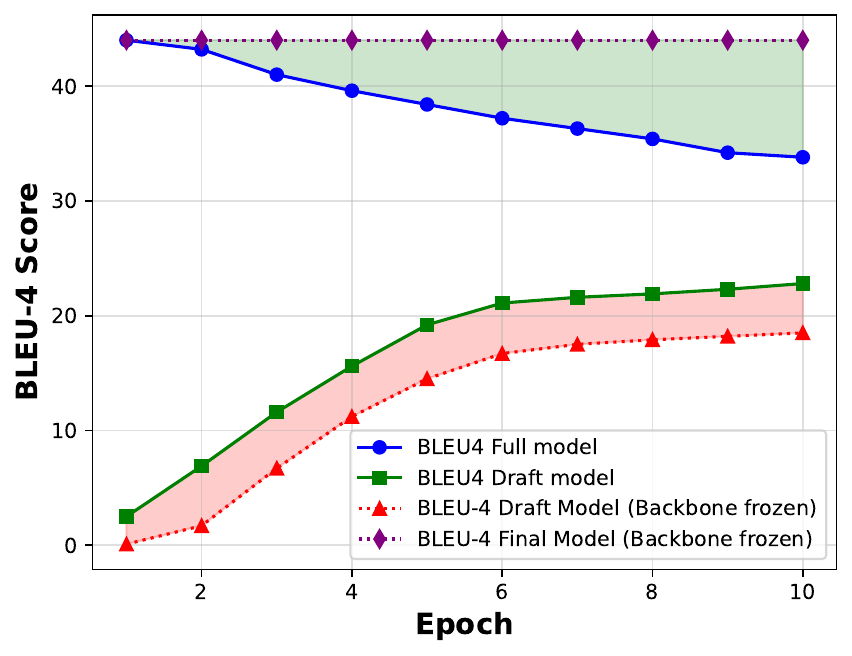}
    \caption{The figure shows how we preserve the model optimality by freezing the backbone while not losing much in terms of draft model performance.}
    \label{fig:Freezed_training}
\end{figure}

In Figure \ref{fig:Freezed_training}, we show that we save a lot of loss in performance by freezing the backbone. The loss incurred was due to shared parameters of the draft model and full model. Freezing the full model and learning only the imitation network to mimic the final layers maintains the backbone optimality and also has minimal loss in draft model performance while being substantially lower in the number of training parameters i.e., if the full backbone is fine-tuned the training parameters are $146M$ parameters and when the backbone is frozen the number of trainable parameters are down to $39M$. However, due to the deep layer imitation, the loss in draft model performance is quite small. 

\subsection{Some instances of SSD inference}\label{sec:instance_discussion}
\begin{figure*}
    \centering
    \includegraphics[scale = 0.5]{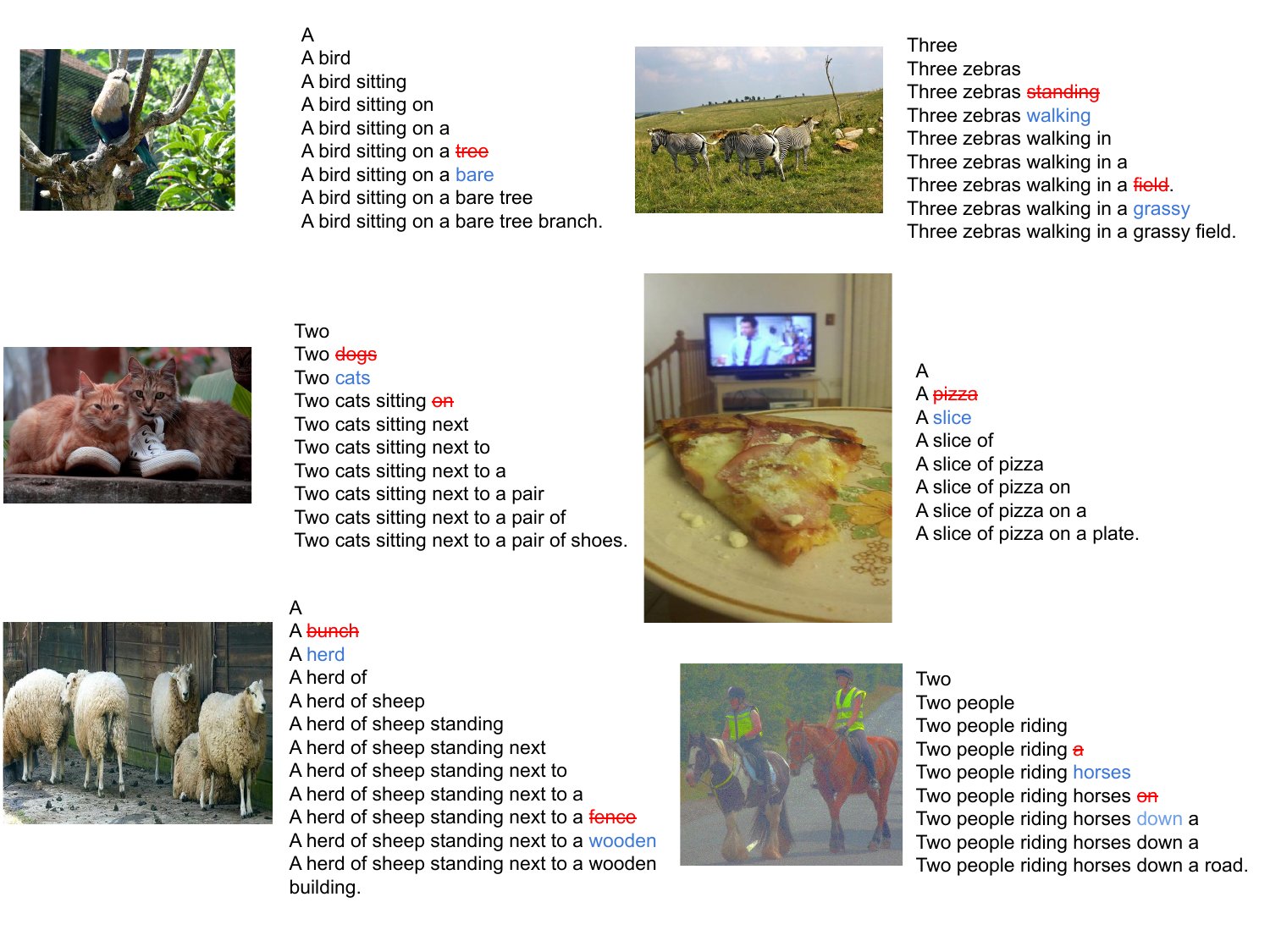}
    \caption{Some instances of the inference performed by our method where struck-out text is the words rejected at the verification step while the text in blue colour is the replacement by the full model. The struck-out text denotes the rejected tokens and the tokens in blue colour denote the corrected token by the full model while the remaining tokens are the accepted ones.}
    \label{fig:more_instances}
\end{figure*}

In Figure \ref{fig:more_instances}, we show the multiple inference scenarios where SSD speeds up the inference process. Observe that the draft model is not able to correctly generate the very fine-grained features of the image. For example, see the first image on the left side of the figure, it shows that the draft model correctly predicts that it is a tree but the full model corrects it by predicting a `bare tree branch' instead of the tree. Once the full model rejects the `tree' prediction of the draft model and corrects it with bare, the draft model then correctly predicts `tree branch ' after `bare'. We observe that the draft model has some difficulty in generating rarely occurring words such as `bare', `grassy' etc. Still, once the full model makes it correct, it starts predicting the correct tokens to complete the captions. Sometimes, due to smaller architectures, the draft model predicts common words such as `bunch' in the third figure on the left side then the full model corrects it by replacing it with herd after which the draft model takes over. For longer context lengths, we use the instance of the COCO dataset image and ask it to explain briefly about the image. Observe that as the context length increases, the rejection of a token was quite small, motivating us to choose a higher value of draft token generation $d$ before a verification step.

\subsection{Runtime}\label{sec:GPUs} 
For conducting the experiments, we used a set of 5 NVIDIA A6000 (48 GB) and 3 NVIDIA RTX 1080 (12 GB) GPUs. The highest training time observed was on the BLIP-2 model for fine-tuning the backbone on the COCO dataset, requiring $7$ hours of runtime for $5$ epochs. During the inference phase, the runtime was less than an hour for all the datasets.

\section{Imitation network architecture}\label{sec:imitation_network}

The imitation network can adopt architectures of varying depth and width, provided its computational overhead remains lower than the efficiency it introduces. Essentially, its complexity must be significantly less than that of the subsequent layers in the backbone following the chosen value of $n$. The search space for such networks includes configurations with diverse layer compositions, such as combinations of dense layers and self-attention mechanisms. However, all architectures within this space are required to produce an output that aligns with the backbone’s output distribution, ensuring the same number of classes.

In our experiments, the search space was restricted to architectures comprising up to two linear layers and a single self-attention layer with a fixed hidden size (same as the hidden size of the model). These components mirrored the structural elements of the decoder model. While the dropout rate was left as a hyperparameter to be optimized, the search process allows customization based on specific requirements or constraints, such as computational budgets.

Several tools exist for performing Neural Architecture Search (NAS), including Auto-Keras \cite{jin2019auto}, Auto-PyTorch \cite{zimmer2021auto}, and Neural Network Intelligence (NNI) \cite{nni}. In line with \cite{khademsohiselfxit}, we utilized NNI to conduct the NAS process, customizing the evaluation to prioritize both model accuracy and computational complexity. Computational complexity was assessed using floating-point operation (FLOPs) counts. However, the NAS process proved to be computationally intensive. Note that, as the input representations vary across different $n$ values, the optimal architecture identified by NAS may also vary. Thus, selecting an appropriate $n$ is a prerequisite.


We found the best architecture for the COCO dataset and then fixed it to other datasets without computing it for every dataset. This is a common practice where an optimal structure is found on a dataset and then fixed across various datasets \cite{geifman2019deep, cui2019fast, elsken2019neural}. This process required $15$ hours of GPU runtime with the same set of GPUs given in section \ref{sec:GPUs} (faster than manual checking). The NAS process identified the optimal design as a stack of two linear layers. This is intuitive as the attention mechanisms are not meant for mimicking scenarios while linear layers can perform better when the task comes to replicate some behaviour. Also, they have reduced computational complexity.

\section{Mixed Training Setting}
\begin{figure}
    \centering
    \includegraphics[scale=0.5]{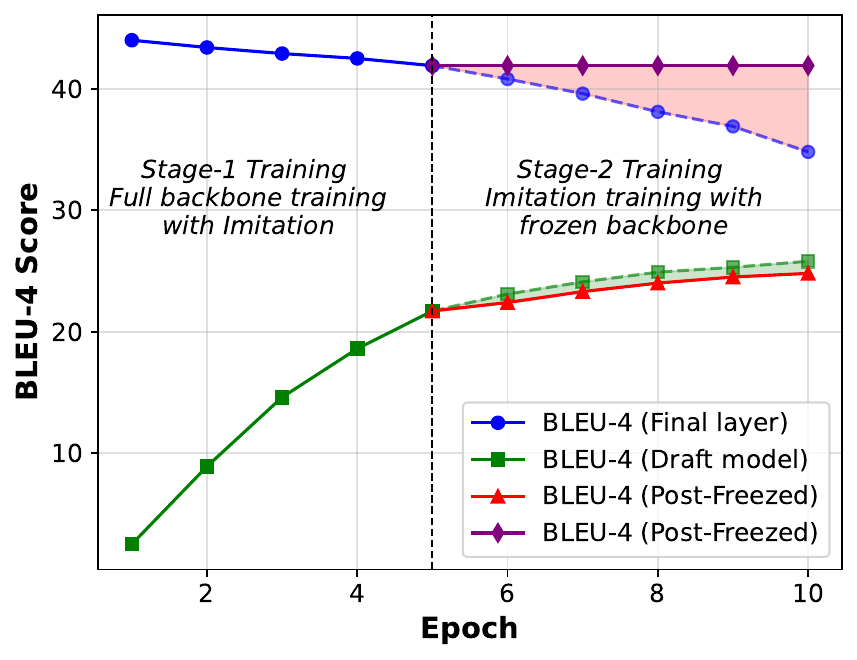}
    \caption{Mixed training where the model is simultaneously trained first and then backbone parameters are frozen and only imitation network is trained.}
    \label{fig:mixed_training}
\end{figure}

We have explored two training setups: 1) where the full model and the draft model with imitation are trained simultaneously and 2) where the full model and the imitation network are trained separately. In figure \ref{fig:mixed_training}, we show the final layer and draft model with imitation performance when the model is initially trained simultaneously for $5$ epochs and then the backbone is frozen while only training the imitation network for later epochs on the BLIP-2-Flant5-xl model. This method can provide a better speedup but with overall performance loss, where the loss will be due to a hit in full model performance caused by simultaneous training. Still, this method is important because it has a higher speedup, as the performance of the draft model with imitation has been improved. The reason for the better model performance of the draft model is that now the imitation network gets help from the Q-former of the BLIP-2 model, which now generates representations that are better for both the draft model as well as final layer. However, as the loss is performance is not that high, the user can treat the number of epochs for unfrozen training as a hyperparameter and achieve the desired speedup with a slight loss in performance.

\begin{table}[]
\centering
\begin{tabular}{l|cl}
\hline
LLM                 & \multicolumn{1}{l}{FlanT5-xl} & OPT-2.7B \\ \hline
Finetuning epochs  & \multicolumn{2}{c}{5}                   \\
Finetuning dataset  & \multicolumn{2}{c}{Train split}                   \\
Imitation epochs  & \multicolumn{2}{c}{20}                   \\
Warmup steps        & \multicolumn{2}{c}{1000}                 \\
Learning rate       & \multicolumn{2}{c}{1e-5}             \\
Batch size          & \multicolumn{2}{c}{16}                   \\
AdamW beta          & \multicolumn{2}{c}{(0.9, 0.999)}         \\
Weight decay        & \multicolumn{2}{c}{0.05}                 \\
Drop path           & \multicolumn{2}{c}{0}                    \\
Image resolution    & \multicolumn{2}{c}{364}                  \\
Prompt              & \multicolumn{2}{c}{``a photo of"}        \\
Inference beam size & \multicolumn{2}{c}{5}                    \\ 
d    &      \multicolumn{2}{c}{8}\\
Fusion-strategy      &    \multicolumn{2}{c}{Concatenation}   \\
\hline
\end{tabular}
\caption{Hyperparameters for training the BLIP2 backbone on image captioning datasets.}
\label{tab: training_hyper_1}
\end{table}

\begin{table}[]
\centering
\begin{tabular}{l|cl}
\hline
LLM                 & LLaMA-2 \\ \hline
Imitation training epochs  & \multicolumn{2}{c}{20}                   \\
Dataset-used  & \multicolumn{2}{c}{Instructllava150k}                   \\
Warmup steps        & \multicolumn{2}{c}{1000}                 \\
Learning rate       & \multicolumn{2}{c}{2e-5}             \\
Batch size          & \multicolumn{2}{c}{16}                   \\
AdamW beta          & \multicolumn{2}{c}{(0.9, 0.999)}         \\
Weight decay        & \multicolumn{2}{c}{0.00}                 \\
Image resolution    & \multicolumn{2}{c}{490}                  \\
Inference beam size & \multicolumn{2}{c}{5} \\  
d    &      \multicolumn{2}{c}{8}\\
n    &      \multicolumn{2}{c}{16}\\
Fusion-strategy      &    \multicolumn{2}{c}{Concatenation}   \\
 \hline
\end{tabular}
\caption{Hyperparameters for training the LLaVA backbone on Instructllava-150k dataset.}
\label{tab: training_hyper_3}
\end{table}

\begin{table}[]
\centering
\begin{tabular}{lccc}
\hline
Model           & \multicolumn{2}{c}{VQAv2} & Speed \\
                & dev         & test        &       \\ \hline
BLIP2-FlanT5-xl & 84.2        & 84.0          & 1.00$\times$     \\
Draft w/o imi   & 09.3         & 08.5         & 2.00$\times$     \\
Draft w imi     & 41.9        & 40.7        & 2.00 $\times$     \\
Draft \& Verify & 73.5        & 71.8        & 1.29$\times$  \\
LayerSkip       & 75.8        & 75.0          & 1.33$\times$  \\ \hline
Our             & 83.9        & 83.4        & 1.39$\times$  \\ \hline
\end{tabular}
\caption{Results on the VQA dataset.}
\label{tab: vqa_res}
\end{table}

\begin{table*}[]
\centering
\begin{tabular}{l|cl}
\hline
LLM                 & \multicolumn{1}{l}{FlanT5-xl} & OPT-2.7B \\ \hline
Finetuning epochs  & \multicolumn{2}{c}{5}                   \\
Finetuning dataset  & \multicolumn{2}{c}{Train split}                   \\
Imitation training epochs  & \multicolumn{2}{c}{20}                   \\
Warmup steps        & \multicolumn{2}{c}{1000}                 \\
Learning rate       & \multicolumn{2}{c}{1e-5}             \\
Batch size          & \multicolumn{2}{c}{16}                   \\
AdamW beta          & \multicolumn{2}{c}{(0.9, 0.999)}         \\
Weight decay        & \multicolumn{2}{c}{0.05}                 \\
Drop path           & \multicolumn{2}{c}{0}                    \\
Image resolution    & \multicolumn{2}{c}{490}                  \\
Prompt              & \multicolumn{2}{c}{``Question:\{\}  Answer:"}        \\
Inference beam size & \multicolumn{2}{c}{5} \\  
d    &      \multicolumn{2}{c}{5}\\
Fusion-strategy      &    \multicolumn{2}{c}{Concatenation}   \\
 \hline
\end{tabular}
\caption{Hyperparameters for training the BLIP2 backbone on VQA datasets.}
\label{tab: training_hyper_2}
\end{table*}